%% file: root.tex
\documentclass[letterpaper, 10 pt, journal]{IEEEtran}  

\IEEEoverridecommandlockouts                              

\usepackage{cite}
\usepackage{amsmath,amssymb,amsfonts}
\usepackage{algorithm}
\usepackage[noend]{algpseudocode}
\usepackage{subcaption}
\usepackage{graphicx}
\usepackage{physics}
\usepackage{textcomp}
\usepackage{hyperref}
\usepackage{gensymb}
\usepackage{multirow}
\let\labelindent\relax
\usepackage{enumitem}
\usepackage[font=small,labelfont=bf]{caption}
\usepackage[table,xcdraw]{xcolor}
\usepackage{adjustbox}
\usepackage{dsfont}
\usepackage{booktabs}
\usepackage{siunitx}
\usepackage{subfiles}

\def\figref#1{Fig.~\ref{#1}}
\def\tabref#1{Tab.~\ref{#1}}
\def\eqref#1{Eq.~(\ref{#1})}

\input{notation.tex}

\input{glossary}

\title{\LARGE \bf
RoadRunner M\&M - Learning Multi-range Multi-resolution  \\ Traversability Maps for Autonomous Off-road Navigation
}

\author{Manthan Patel$^{1,2}$, Jonas Frey$^{1,2}$, Deegan Atha $^{1}$, Patrick Spieler $^{1}$, Marco Hutter $^{2}$ and Shehryar Khattak $^{1}$ 
\thanks{$^1$Jet Propulsion Laboratory (JPL), California Institute of Technology(Caltech), Pasadena, CA, United States of America}
\thanks{$^2$Swiss Federal Institute of Technology (ETH Zurich), Robotic Systems Lab, Switzerland}\\
\thanks{The research was carried out at the Jet Propulsion Laboratory, California Institute of Technology, under a contract with the National Aeronautics and Space Administration (80NM0018D0004). This work was partially supported by Defense Advanced Research Projects Agency (DARPA).}
\thanks{\copyright 2024. California Institute of Technology. Government sponsorship acknowledged. All rights reserved.}
}

\begin{document}
\bstctlcite{IEEEexample:BSTcontrol}

\maketitle
\thispagestyle{empty}
\pagestyle{empty}

\subfile{00_abstract}

\subfile{01_introduction}

\subfile{02_related}

\subfile{03_approach}

\subfile{04_experiments}

\subfile{05_conclusion}



\bibliographystyle{IEEEtran}
\bibliography{ref}


\end{document}

%% file: notation.tex
\newcommand{\pose}[3]{\mathbf{T}_{\mathtt{#1 #2}}^{#3}}

\newcommand{\rot}[3]{\mathbf{R}_{\mathtt{#1 #2}}^{#3}}
\newcommand{\pos}[2]{{\mathtt{_#1}} \mathbf{p}^{#2}}

\newcommand{\K}[1]{\leftindex{^#1}\mathbf{K}_{3\times3}}
\newcommand{\img}[3]{\leftindex{^#1}\mathbf{I}_{#2}^{#3}}

\newcommand{\feat}[4]{\mathbf{F}_{#1 \times #2 \times #3}^{#4}}

\newcommand{\grid}[2]{\mathbf{G}_{#1}^{#2}}
\newcommand{\gridhat}[2]{\mathbf{\hat{G}}_{#1}^{#2}}

\newcommand{\loss}[1]{\mathcal{L}_{\mathrm{#1}}}

\newcommand{\fun}[2]{f_{\mathrm{#1}}\left( #2 \right) }

\robustify{\pose}
\robustify{\rot}
\robustify{\pos}
\robustify{\K}
\robustify{\loss}
\robustify{\feat}
\robustify{\img}
\robustify{\fun}

\usepackage{xspace}
\makeatletter
\DeclareRobustCommand\onedot{\futurelet\@let@token\@onedot}
\def\@onedot{\ifx\@let@token.\else.\null\fi\xspace}
\makeatother
\newcommand{\eg}{e.g\onedot}

\newcommand{\short}{\textit{short}\xspace}
\newcommand{\micr}{\textit{micro}\xspace}
\newcommand{\ls}{\textit{Lift Splat}\xspace}
\newcommand{\pp}{\textit{PointPillars}\xspace}

\newcommand{\obsf}{Obs.~F~({\color[HTML]{962491}$\blacksquare$})\xspace}
\newcommand{\obsp}{Obs.~PC~({\color[HTML]{fb9727}$\blacksquare$})\xspace}
\newcommand{\unobs}{Unobs.~({\color[HTML]{436eb0}$\blacksquare$})\xspace}

\newcommand{\obspt}{{\color[HTML]{fb9727}Obs.~PC}\xspace}
\newcommand{\obsft}{{\color[HTML]{962491}Obs.~F}\xspace}
\newcommand{\unobst}{{\color[HTML]{436eb0}Unobs.}\xspace}

\newcommand{\unobssq}{{\color[HTML]{436eb0}$\blacksquare$}\xspace}
\newcommand{\obssq}{{\color[HTML]{fb9727}$\blacksquare$}{\color[HTML]{962491}$\blacksquare$}\xspace}

\newcommand{\magenta}[1]{\color[HTML]{962491}{#1}}
\newcommand{\orange}[1]{\color[HTML]{fb9727}{#1}}
\newcommand{\blue}[1]{\color[HTML]{436eb0}{#1}}
\newcommand{\grey}[1]{\color[HTML]{656565}{#1}}

%% file: glossary.tex
\usepackage[abbreviations]{glossaries-extra}

\glssetcategoryattribute{abbreviation}{indexonlyfirst}{true}

\glssetcategoryattribute{abbreviation}{nohyperfirst}{true}

\newabbreviation{auroc}{AUROC}{Area Under the Receiver Operating Characteristic Curve}
\newabbreviation{accuracy}{Acc}{Accuracy}

\newabbreviation{bev}{BEV}{Bird`s Eye View}
\newabbreviation{camera}{}{Carnegie Robotics MultiSense S27}
\newabbreviation{cnn}{CNN}{Convolutional Neural Network}
\newabbreviation{cvar}{CVaR}{Conditional Value at Risk}

\newabbreviation{dem}{DEM}{Digital Elevation Map}

\newabbreviation{fov}{FoV}{Field of View}
\newabbreviation{fpn}{FPN}{Feature Pyramid Network}
\newabbreviation{gnn}{GNN}{Graph Neural Network}
\newabbreviation{gcn}{GCN}{Graph Convolutional Network}
\newabbreviation{gt}{GT}{ground truth}
\newabbreviation{hdd}{HDD}{Hazard Distance Detection}
\newabbreviation{imu}{IMU}{Inertial Measurement Unit}
\newabbreviation{irl}{IRL}{Inverse Reinforcement Learning}
\newabbreviation{icp}{ICP}{Iterative Closest Point}
\newabbreviation{knn}{KNN}{K-Nearest Neighbors}

\newabbreviation{lagr}{LAGR}{Learning Applied to Ground Vehicles}
\newabbreviation{lidar}{LiDAR}{Light Detection and Ranging}
\newabbreviation{ladar}{LADAR}{LAser Detection And Ranging}
\newabbreviation{lidarsensor}{}{Velodyne VLP-32C}

\newabbreviation{mlp}{MLP}{Multi-Layer Perceptron}
\newabbreviation{mpc}{MPC}{Model Predictive Controller}
\newabbreviation{mdp}{MDP}{Markov Decision Process}
\newabbreviation{mse}{MSE}{Mean Squared Error}
\newabbreviation{mae}{MAE}{Mean Absolute Error}
\newabbreviation{mppi}{MPPI}{Model Predictive Path Integral}

\newabbreviation{nir}{NIR}{Near-InfraRed}

\newabbreviation{wmse}{WMSE}{Weighted Mean Squared Error}
\newabbreviation{wmae}{WMAE}{Weighted Mean Absolute Error}

\newabbreviation{ours}{RoadRunner M\&M}{RoadRunner M\&M}

\newabbreviation{rbf}{RBF}{Radial Basis Function}
\newabbreviation{rmp}{RMP}{Riemannian Motion Policies}
\newabbreviation{ros}{ROS}{Robot Operating System}
\newabbreviation{ros1}{ROS~1}{Robot Operating System}
\newabbreviation{roc}{ROC}{Receiver Operating Characteristic}
\newabbreviation{rf}{RF}{Random Forest}
\newabbreviation{radar}{RADAR}{RAdio Detection And Ranging}
\newabbreviation{rl}{RL}{
Reinforcement Learning}
\newabbreviation{rr}{RoadRunner}{RoadRunner}

\newabbreviation{sdf}{SDF}{Signed Distance Field}
\newabbreviation{slam}{SLAM}{Simultaneous Localization and Mapping}
\newabbreviation{svm}{SVM}{Support Vector Machine}
\newabbreviation{svc}{SVC}{Support Vector Classifier}
\newabbreviation{stack}{X-Racer}{X-Racer}

\newabbreviation{work}{TraFo}{TerrainFormer}

\newabbreviation{ugv}{UGV}{Unmanned Ground Vehicle}

\newabbreviation{usgs}{USGS}{Unites States Geological Survey}
\newabbreviation{wvn}{WVN}{Wild Visual Navigation}

\newabbreviation{vit}{ViT}{Vision Transformer}
\newabbreviation{vehicle}{}{Polaris RZR S4 1000 Turbo}

%% file: 00_abstract.tex
\begin{abstract}


Autonomous robot navigation in off--road environments requires a comprehensive understanding of the terrain geometry and traversability. The degraded perceptual conditions and sparse geometric information at longer ranges make the problem challenging especially when driving at high speeds. Furthermore, the sensing--to--mapping latency and the look--ahead map range can limit the maximum speed of the vehicle. 
Building on top of the recent work \glsentrylong{rr}, in this work, we address the challenge of long-range ($\pm\SI{100}{m}$) traversability estimation. Our RoadRunner (M\&M) is an end-to-end learning-based framework that directly predicts the traversability and elevation maps at multiple ranges ($\pm\SI{50}{m}$, $\pm\SI{100}{m}$) and resolutions (\SI{0.2}{m}, \SI{0.8}{m}) taking as input multiple images and a LiDAR voxel map. Our method is trained in a self--supervised manner by leveraging the dense supervision signal generated by fusing predictions from an existing traversability estimation stack (\glsentrylong{stack}) in hindsight and satellite Digital Elevation Maps. \glsentrylong{ours} achieves a significant improvement of up to 50\% for elevation mapping and 30\% for traversability estimation over \glsentrylong{rr}, and is able to predict in 30\% more regions compared to \glsentrylong{stack} while achieving real--time performance. Experiments on various out--of--distribution datasets also demonstrate that our data-driven approach starts to generalize to novel unstructured environments. We integrate our proposed framework in closed--loop with the path planner to demonstrate autonomous high--speed off--road robotic navigation in challenging real--world environments. \textit{Project Page--\url{https://leggedrobotics.github.io/roadrunner_mm/}}

\end{abstract}

%% file: 01_introduction.tex
\section{Introduction}

 \begin{figure}
    \centering
    \includegraphics[width=\columnwidth]{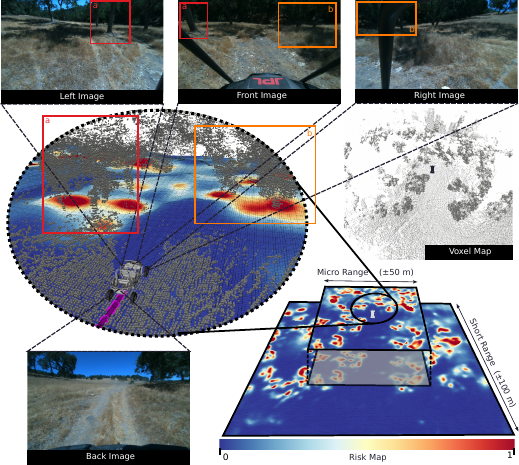}
    \caption{\glsentrylong{ours} takes as input four RGB images and a LiDAR voxel map to predict traversability (risk) and elevation maps at multiple ranges: high resolution \micr range ($\pm\SI{50}{m}$) and low resolution \short range ($\pm\SI{100}{m}$). In the above example, the vehicle is traversing through a dense forest environment. In the zoomed-in version of the \micr range risk map, the risk associated with the trees (a, b) can be clearly visualized.
    }
    \label{fig:cover_pic}
    \vspace{-4ex}
\end{figure}

Autonomous robotic navigation in challenging off--road environments has diverse critical applications, including search-and-rescue missions, planetary exploration, environmental monitoring, and agriculture. To navigate safely, a reliable assessment of terrain traversability is crucial. This is particularly difficult for off--road environments as, unlike urban environments where roads define traversability, there is no clear distinction between traversable and non-traversable regions. Furthermore, the unavailability of prior maps, unreliable GPS, and the presence of obscurants, such as dust, fog, and rain, add to the challenges of off--road robotic navigation. \par

For safe high--speed off-road driving, obtaining precise traversability predictions at a low latency, which reflect potential hazards at long distances, is critical.  
In this work, we define long-range as distances of $\pm\SI{100}{m}$, where partial observations—caused by occlusions, limited sensor coverage, and sparse geometric information—make heuristic-based approaches impractical and unscalable.
Recently, data-driven approaches address some of the issues~\cite{frey2024roadrunner, meng2023terrainnet, chung2024pixel}, with \glsentrylong{rr}~\cite{frey2024roadrunner} proposing an approach to leverage multiple sensing modalities (image and LiDAR data) to predict terrain traversability and elevation at low latency. 
However, although \glsentrylong{rr} demonstrated promising results, it was only evaluated within the same ecological region and limitations on prediction range and temporal consistency, restricted the reliability required for real--time path planning for real--world operations.
Moreover, it is important to have varying map resolution with range. In close proximity of the robot, higher mapping resolution is needed to capture terrain risks according to robot dynamics and to capture the high frequency elevation changes such as ditches and ruts.
Farther from the vehicle, maps capturing information at a coarser scale but at longer ranges are required to plan smoother paths to facilitate high--speed navigation, for \eg detecting a cluster of trees far away to plan around them instead of reacting when close.

Motivated by the discussion above, this work proposes a learning-based approach for simultaneous prediction of terrain traversability and elevation maps at multiple ranges and resolutions~(\figref{fig:cover_pic}) using an end-to-end network.
Inspired by the multi--modal fusion network of~\cite{frey2024roadrunner}, this works builds upon the \glsentrylong{rr} architecture and introduces several components including a novel multi-range multi-resolution hierarchical decoder, LiDAR voxel map input and satellite \gls{dem} for dense supervision signal, which significantly improve the performance while reducing the latency.

The main contributions of the proposed work are as follows:
\begin{itemize}[noitemsep]
    \item \glsentrylong{ours} (Multi-range and Multi-resolution), a novel end-to-end network for simultaneously predicting elevation maps and traversability maps at multiple ranges and resolutions at low latency.
    \item Evaluation on real--world datasets with up to 50\% improvement for elevation mapping and 30\% for traversability estimation, over \glsentrylong{rr}, while providing 30\% more map coverage over \glsentrylong{stack}.
    \item Evaluations for zero-shot deployments in various ecologically distinct out-of-distribution environments, including a desert, beach, canyon, and dense forest.
    \item Demonstration of real--world high-speed field experiments by integrating \glsentrylong{ours} within a full autonomous off-road navigation stack.
\end{itemize}

%% file: 02_related.tex
\section{Related Work}
\subsection{On-Road BEV Map Learning}
The \gls{bev} map representation is widely adopted in autonomous driving and mobile robotics due to its compatibility with downstream tasks and ability to fuse multi--modal sensor data.
For incorporating image features, the forward projection method was pioneered by Lift Splat Shoot~\cite{philion2020lift}, where a per-pixel predicted depth distribution is used to \textit{lift} the image feature into 3D space and then \textit{splat} into a top-down \gls{bev} grid. Differently, in backward projection, a predefined 3D grid \textit{pulls} the image features onto the 3D grid~\cite{simplebev, li2022bevformer, detr3d}. Recent work FB-BEV \cite{li2023fbbev} combines both forward and backward projections to enable effective transformations. Another advantage of using a \gls{bev} map representation is that it allows to fuse different sensing modalities such as LiDAR \cite{liu2022bevfusion, liang2022bevfusion, gunn2023liftattendsplat} and Radars \cite{simplebev}. \glsentrylong{ours} uses a similar fusion strategy of \cite{liu2022bevfusion} and forward projection method of \cite{philion2020lift}.
\subsection{Off-Road Traversability Learning}
In \cite{semantic_offorad_maturana}, a CNN extracts semantic features from images, which are projected onto a 2.5D map using the LiDAR point clouds, yielding a 2.5D semantic map. \cite{schilling_terrain} fit a random forest classifier on a semantic image and geometric LiDAR features to classify terrain in fixed traversability classes. In \cite{locotravfrey}, a 3D voxel map is used to predict the traversability while making use of parallelization in simulation to generate the supervision signal. \cite{ruetz2023foresttrav} also uses a voxel map input with sparse 3D CNN to predict traversability but utilizes hand-labelled ground truth traversability maps for supervision. BADGR~\cite{badgr} predicts future events such as collision and terrain properties to train a policy to avoid collisions and prefer smooth terrains. In WayFAST\cite{wayfast}, traction estimates provided by an online receding horizon estimator are used as a proxy for the traversability supervision signal for terrain traversability. Wild Visual Navigation \cite{frey23fast} leverages pre-trained image features to adapt a traversability estimation model online during deployment using a velocity-tracking criterion. \cite{schmid_trav} predict the traversability using the reconstruction error of an autoencoder trained using human driving data. V-STRONG \cite{jung2024vstrong} employs contrastive representation learning using both human driving data and instance segmentation from a vision foundation model as the supervision signal for predicting traversability. In \cite{irl_costmap}, inverse reinforcement learning is used to learn risk-aware costmaps leveraging a fast \gls{mpc} approach for solving the \gls{mdp}. EVORA \cite{cai2023evora} presents a framework to learn an uncertainty-aware traction model and plans risk-aware trajectories.
\subsection{Off-Road BEV Map Learning}
In \cite{shaban22a_bevnet}, authors introduced a sparse 3D CNN operating on LiDAR point clouds to classify the terrain into fixed traversability classes in \gls{bev} space. TerrainNet~\cite{meng2023terrainnet} introduced a framework for semantic segmentation and elevation mapping in \gls{bev} space, demonstrating that using stereo depth and RGB images leads to accurate predictions. However, the prediction range is limited to $\pm\SI{25}{m}$, where the stereo depth is reliable. Pixel-to-elevation\cite{chung2024pixel} introduces a cross-view transformer-based architecture to perform long range elevation mapping while making use of the satellite \gls{dem}s as the supervision signal. In WayFASTER \cite{gasparino2024wayfaster}, the self-supervision concept of WayFAST \cite{wayfast} is extended to \gls{bev} space along with temporal fusion and depth inputs for improved performance. However, the supervision signal is sparse and requires a traction model in combination with accurate state estimation. Recently, \cite{aich2024deep} present an approach for inpainting high resolution \gls{bev} maps by leveraging a generative model formulation. In \glsentrylong{rr}~\cite{frey2024roadrunner}, the authors introduced a multi--modal network taking as input RGB images and LiDAR point clouds to predict elevation and traversability maps. 

%% file: 03_approach.tex
\section{Methodology}

\subsection{Problem Statement}
\label{sec:problem}
Our objective is to predict elevation maps $\grid{ele}{\rho}\in{\mathbb{R}}^{{H_\rho/r_\rho}\times {W_\rho/r_\rho}\times 1}$ and traversability maps $\grid{trav}{\rho}\in{\mathbb{R}}^{{H_\rho/r_\rho}\times{W_\rho/r_\rho}\times 1}$ at two different ranges and resolutions (\figref{fig:cover_pic}) in a vehicle-centric gravity-aligned frame. The center of these grid maps is defined by the position and yaw orientation of the vehicle. Following terminology of~\cite{frey2024roadrunner}, we define the ranges $\rho \in \{m:micro, s:short\}$, where $(H_{m},W_{m},r_{m})=$ (\SI{100}{m}, \SI{100}{m}, \SI{0.2}{m}) and $(H_{s},W_{s},r_{s})=$ (\SI{200}{m}, \SI{200}{m}, \SI{0.8}{m}). Hence, the \micr range maps $\grid{}{m}$ have vehicle--centered range of $\pm$\SI{50}{m} at a higher resolution of \SI{0.2}{m} and the \short range maps ${\bf{G}}^{s}$ have a range of $\pm$\SI{100}{m} at a lower resolution of \SI{0.8}{m}. 

\subsection{X-Racer Overview}
\label{sec:xracer}
We leverage NASA Jet Propulsion Laboratory's off--road autonomy research stack \glsentrylong{stack}~(See Sec. 3.3 of \cite{frey2024roadrunner} for more details) for generating training labels. For our experiments, the \glsentrylong{stack} has been deployed on a modified Polaris RZR all-terrain vehicle. Four MultiSense S27  
(front, left, right, and back) cameras provide RGB images, and point cloud data is obtained from the three Velodyne VLP-32C LiDARs 
(front, front-tilted, and back). All of the sensors are hardware time-synchronized. The vehicle is equipped with a Threadripper 3990x CPU and 4xGeForce RTX 3080 GPUs. 
Semantic segmentation is performed on the input images using Segmenter\cite{strudel2021segmenter} (Sec. 3.3.2 of \cite{frey2024roadrunner}) and then projected onto LiDAR points to obtain a semantic point cloud, which is temporally aggregated to obtain a vehicle-centric voxel map (Sec. 3.3.3 of \cite{frey2024roadrunner}). This is performed for both \micr and \short ranges. Traversability and elevation maps are then derived from the voxel maps using heuristics tuned in simulation and on real--world data (Sec. 3.3.4 of \cite{frey2024roadrunner}). The traversability risk value of 0 and 1 indicate safe and unsafe, respectively. Additionally, a confidence map is also generated which accounts for the density of LiDAR points and the minimum vehicle distance to voxel.
The selection of \micr range resolution of \SI{0.2}{m} is driven by the tire width of our vehicle, while the \short range resolution of \SI{0.8}{m} is chosen for computational efficiency. 
While the \glsentrylong{stack} facilitates safe and autonomous off-road navigation, it can be unreliable in regions with sparse geometric data, particularly at longer ranges and at higher speeds due to limited LiDAR update rates and sparse returns. Additionally, the \glsentrylong{stack} naively interpolates or extrapolates in regions with missing geometric information, which we argue can be better predicted using information from the images. 
Lastly, \glsentrylong{stack}'s multi-step map generation process introduces a significant latency~(\SI{500}{ms}) from sensor data to traversability estimation, which constrains the safe speed limit.

\subsection{Pseudo Ground Truth Generation}
\label{sec:gtgen}

\begin{figure}
    \centering
    \includegraphics[width=\columnwidth]{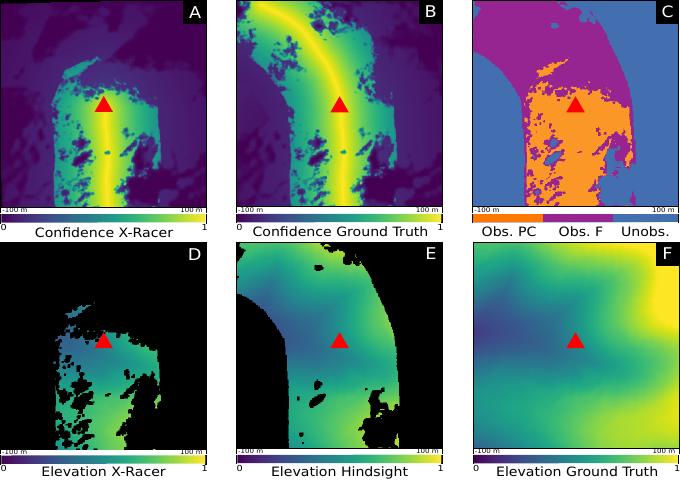}
    \caption{The vehicle is traversing up a hill. The red triangle represents the pose of the vehicle. Various \short range maps are visualized. The \glsentrylong{stack} stack is able to confidently (A) predict the elevation maps only in vehicle proximity (D) where geometric observations are available. By accumulating the future predictions in hindsight, we generate the accurate ground truths (B, E) in the regions traversed by the car in future. Complete ground truth maps are generated by fusing the USGS DEMs (F). (E) represents the regions as observed in past and current observations (\obsp), Future observations (\obsf) and unobserved regions (\unobs).
    }
    \label{fig:dem}
    \vspace{0ex}
\end{figure}

\begin{figure*}[ht]
    \centering
    \includegraphics[width=\textwidth]{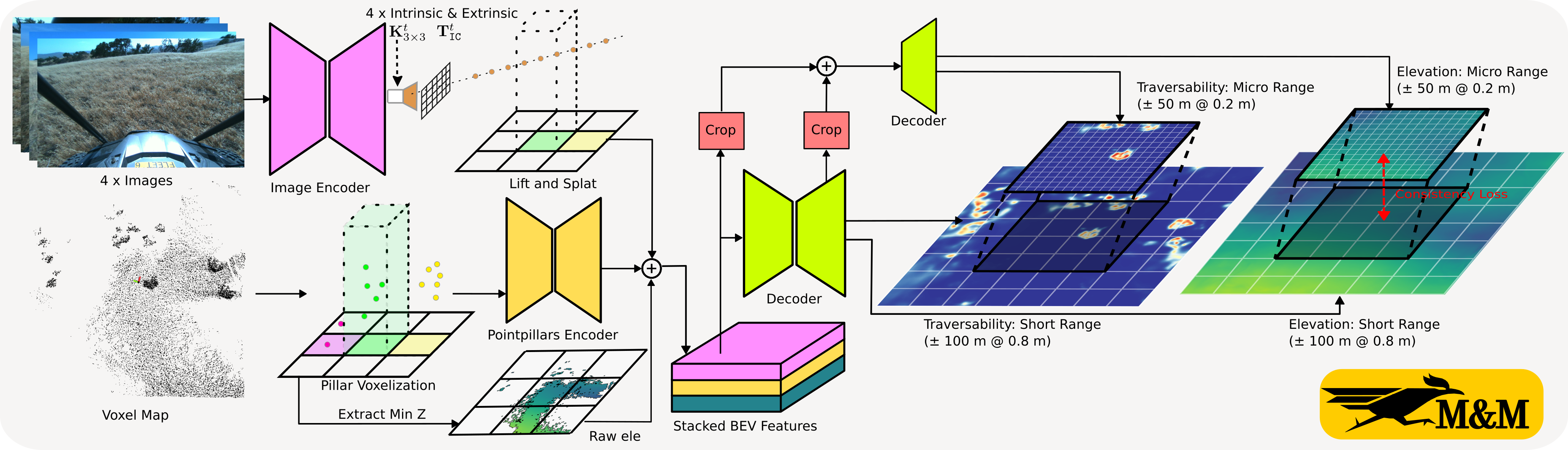}
    \caption{Overview of the \glsentrylong{ours} network architecture. The network takes as an input four RGB images which are encoded using the \ls method~\cite{philion2020lift}. \pp~\cite{Lang2018PointPillarsFE} encoding is used for the input voxel map. Additonally, a raw elevation map is extracted from the voxel map using the min Z values. These multi-modal features are stacked and passed through a hierarchical decoder which predicts the maps at different ranges and resolutions.
    }
    \label{fig:pipeline}
    \vspace{-4ex}
\end{figure*}

In the following, we use the terms ground truths and "pseudo" ground truths interchangeably since it is a reasonable proxy for the actual ground truth in the scope of this work. We adopt an approach similar to \cite{frey2024roadrunner}, leveraging the \glsentrylong{stack} stack and employing hindsight fusion to generate the pseudo ground truth labels for training. The \glsentrylong{stack} predictions are accumulated over time, improving traversability and elevation maps beyond the current reliable perception range, and are used as ground truth during training. Elevation fusion employs a cell-wise mean (empirically found to be able to handle minor Z odometry drifts), confidence fusion uses a cell-wise maximum, and traversability fusion incorporates the latest measurement alongside a confidence threshold. The latest measurements for traversability are used given that the predictions of \glsentrylong{stack} improve over time as more information is accumulated. We use a \SI{60}{s} accumulation time, which proved sufficient for populating \short range maps while having minimal odometry drift. To improve the elevation ground truth map coverage, particularly in the \short range maps lateral to vehicle's path (\figref{fig:dem}), we leverage \gls{dem} obtained from the \gls{usgs}. The \gls{dem}s available at a resolution of 1 meter, are upsampled using bilinear interpolation and used to inpaint the missing elevation values.
Initial alignment between the queried \gls{dem} and hindsight-generated ground truth maps is performed based on the vehicle's GNSS data. However, as GNSS alignment may not be perfect, therefore \gls{icp} registration is employed to refine alignment. Samples with fitness values less than a specified threshold are rejected to ensure high quality elevation. 

\subsection{Network Architecture}
\label{sec:architecture}
An overview of the \glsentrylong{ours} architecture is presented in \figref{fig:pipeline}. The network uses the \ls method~\cite{philion2020lift} and the \pp method ~\cite{Lang2018PointPillarsFE} for encoding image data and LiDAR voxel map data, respectively. All the \gls{bev} features are then fused and passed through our hierarchical decoder, which predicts the traversability and elevation maps at required resolutions and ranges.
For ground truth elevation maps, we limit the height difference to $\pm\SI{25}{m}$ and accordingly rescale them to a range of $\pm1$.

\subsubsection{Image BEV Features}
We use an EfficientNet-B0~\cite{effnet} backbone, shared across all four camera images to obtain multi-scale image features. These are then passed through a \gls{fpn} to fuse the multi-scale features to obtain the per-pixel discrete depth distribution along with the pixel features. The pixel features are lifted along the camera ray using the depth distribution, camera intrinsics and extrinsics. The resulting camera feature point cloud is splat into a \gls{bev} feature grid using an efficient \gls{bev} pooling~\cite{liu2022bevfusion}.

\subsubsection{Point cloud BEV Features}
LiDAR scans are sparse at longer ranges, especially when the vehicle is traversing at higher speeds. To mitigate this, we employ a voxel map to temporally aggregate LiDAR scans, and input the map to the \pp method backbone~\cite{Lang2018PointPillarsFE}.  The input voxel map is discretized into pillars, and additional statistics and features per-pillar are computed. 
A simplified version of PointNet (as in \cite{Lang2018PointPillarsFE}) is applied to process the pillars, which provide higher dimensional features per pillar. The obtained feature grid is then processed by a 2D CNN backbone to obtain the point cloud features in \gls{bev} space.


\subsubsection{Multi-Modal Fusion}
Extracting the height of the lowest occupied voxel along the z-direction can already provide a good prior for the elevation map. We thus stack this additional channel of raw elevation information along with the image and point cloud \gls{bev} features to obtain the multi-modal \gls{bev} features, which are then processed by the hierarchical decoder.

\subsubsection{Hierarchical Multi--resolution Decoder}
We use a shared decoder for traversability and elevation maps as it provides a good trade-off between speed and performance. The hierarchical decoder adopts a U--Net structure with multiple residual blocks and generates the \short range feature maps, which are passed through $1\times1$ convolutional layer to produce the \short range traversability and elevation maps. The multi--modal and the \short range feature maps are then center--cropped within a range of $\pm\SI{50}{m}$, concatenated, and passed through upsampling and convolutional blocks to predict the \micr range maps. 

\subsubsection{Loss Functions}
We denote the predicted gridmaps as $\gridhat{}{}$ and the ground truth gridmaps as $\grid{}{}$. We employ the \gls{mse} loss for traversability ($\mathcal{L}_{\mathrm{trav}}^{\rho}$). For elevation, we use the Smooth-L1 loss and apply different weighting for observed~($\bf{G_O}$\obssq) and unobserved~($\bf{G_U}$\unobssq) regions (\figref{fig:dem}C). Grid cells with a ground truth confidence value greater than 0.1 are considered as observed (\figref{fig:dem}E) and the rest, unobserved. Predicting in unobserved regions is challenging since they are occluded and lack geometric information. Hence, a lower weight is assigned to mitigate their negative impact on training. Thus, the elevation loss is: 

\vspace{-5mm}
\begin{multline}
    \mathcal{L}_{\mathrm{ele}}^{\rho} = \frac{1}{|\grid{o}{}|}\sum_{x,y\in \bf{G_o}} SmoothL1(\grid{ele}{\rho}(x,y), \gridhat{ele}{\rho}(x,y)) \\
    + \frac{\alpha}{|\grid{u}{}|}\sum_{x,y\in \bf{G_u}} SmoothL1(\grid{ele}{\rho}(x,y), \gridhat{ele}{\rho}(x,y))
\end{multline}
Additionally, we penalize the network with a consistency loss $\mathcal{L}_{\mathrm{cons}}$ if it outputs inconsistent elevation values in the overlapping regions at different ranges. For this, we use a Smooth-L1 loss between the center-cropped \short range elevation map and the downsampled \micr range elevation map. The final loss $\mathcal{L}_{\mathrm{total}}$ is formulated as a weighted combination of the aforementioned losses:
\begin{equation}
    \mathcal{L}_{\mathrm{total}} = \sum_{\rho\in\{m, s\}}(\mu\mathcal{L}_{\mathrm{trav}}^{\rho} + \lambda\mathcal{L}_{\mathrm{ele}}^{\rho}) + \gamma\mathcal{L}_{\mathrm{cons}}.
\end{equation}

\subsubsection{Implementation Details}

We use pre-trained weights from ImageNet for EfficientNet-B0. The network takes rectified, downsampled, and normalized images of resolution $396 \times 640$, and the multiscale features at \{5,6,8\} stages are passed to the image \gls{fpn}. The output of the \gls{fpn} is at a resolution of $1/8$ the original input dimensions. These image features are lifted using depth distribution between \SI{1}{m} to \SI{110}{m} with intervals of \SI{0.8}{m} and splatted into a \gls{bev} feature grid of dimensions $250 \times 250$ (similar to the \short range target grid dimensions) with a channel dimension of 80. The \short range voxel map is used as an input to the \pp encoder after re-voxelizing into pillars of resolution \SI{0.8}{m} in x and y directions for a range of $\pm\SI{100}{m}$. We use a maximum of 16 points for each pillar and a maximum number of pillars equal to $(32000, 64000)$ for training and testing, respectively. The output of the \pp encoder is a \gls{bev} feature grid of size $250 \times 250$ with 256 channels. The output of the hierarchical multi--resolution decoder is of size $500 \times 500$ and $250 \times 250$ for \micr and \short range maps, respectively. 
Overall, our network consists of 19.5 M parameters. For the loss, we use weights $\alpha=0.2, \mu=2, \lambda=2$ and $\gamma=5$.
We train the network for a total of 16,000 optimizer steps using the Adam optimizer~\cite{adam} with a learning rate of $5\text{e-}4$ and OneCyleLearningRate schedule. The network is trained on an Nvidia A100 GPU with a batch size of 6. All hyper-parameters related to network size were selected on the basis of inference time, while the rest were determined using a grid search based on the validation dataset performance.

%% file: 04_experiments.tex
\section{Experiments and Results}
\subsection{Robotic Field Deployments and Datasets}
\label{sec:dataset}
To collect real--world training and test datasets, multiple robotic field deployments were conducted on dry grasslands and rolling hills at Halter Ranch near Paso Robles, CA, USA.
A total of \SI{27}{km} of off--road driving data was collected resulting in around 14.2k samples. The data was processed by the \glsentrylong{stack} stack, hind-sight fusion, and \gls{dem} fusion to generate the ground truth maps. The dataset consists of 14 trajectories, which we split into eight training and six test sequences without a geographic overlap.
Training trajectories are further split into a 80/20 train/validation sets, resulting in 8k training, 2k validation, and 4.2k test samples.
Furthermore, we collected out-of-distribution datasets within a desert, dense forest, beach, and canyon shown in Figs. \ref{fig:cover_pic} and \ref{fig:ood}.

\subsection{Evaluation Metrics and Baselines}
We use the \gls{mae} to evaluate the elevation mapping performance following \cite{frey2024roadrunner, meng2023terrainnet}. Evaluation is conducted across three distinct regions within the complete map: observed in the past and current (\obsp), observed in the future (\obsf), and unobserved (\unobs) (Refer to \figref{fig:dem}). \obsp consists of regions where confident geometric observations are available from past or current observations. These regions can be predicted more reliably since the voxel map will contain the geometric information; however, features such as tall grass pose challenges. Next, \obsf consists of regions that are currently not observable but will become observable in the future since the vehicle will be moving in that direction. These regions are often occluded and contain little to no geometric information. Lastly, \unobs regions are the most difficult to predict as they may have no visual or geometric information (\eg perpendicular to vehicle's path).
For traversability estimation, we evaluate the \gls{mse} performance. Following \cite{frey2024roadrunner}, we also evaluate the hazardous region classification, which is critical for safe operation. We apply a \textit{fatal risk} threshold to classify the predictions and ground truth into hazardous and safe regions, to evaluate: Precision, Recall, and F1-score. \par

We compare the performance of \glsentrylong{ours} with different baselines, namely LSS~\cite{philion2020lift}, \pp~\cite{Lang2018PointPillarsFE}, \glsentrylong{rr}~\cite{frey2024roadrunner} and \glsentrylong{stack} to understand the relative performance improvement. We adapted the above mentioned approaches to elevation and traversability estimation tasks and modified the network architecture to be as similar as ours in terms of components and parameters to ensure a fair comparison. We train separate networks for \micr and \short ranges. Additionally, we also compare our approach without the multi--range setting, to highlight the impact of having a single network to predict at multiple ranges. For all results, we take the average of three runs trained with different random seeds.

\begin{table}[t]
\centering
\caption{Evaluation of Elevation Mapping; \glsentrylong{stack} can only predict partially in (X\%) shown in gray. C: Camera, L: LiDAR, E: raw elev., VM:Voxel Map, MR: Multi-Range}
\label{tab:elevation}
\renewcommand{\arraystretch}{1.3}
\resizebox{\columnwidth}{!}{%
\begin{tabular}{@{}llcllll@{}}
\toprule
\multirow{2}{*}{Method} & \multirow{2}{*}{Input} & \multicolumn{1}{l}{\multirow{2}{*}{}} & \multicolumn{4}{c}{Elevation MAE [m] $\downarrow$}                         \\ \cmidrule(l){4-7} 
                        &                        & \multicolumn{1}{l}{}                  & \obspt        & \obsft         & \unobst         & Total          \\ \midrule
                        &                        & \multirow{7}{*}{\rotatebox{90}{Micro Range}}          & \orange{(33.3 \%)}      & \magenta{(46.4 \%)}      & \blue{(20.3 \%)}      &                \\
LSS                     & C                      &                                       & 0.819          & 1.01           & 2.093          & 1.167          \\
Point Pillars           & L                      &                                       & 0.41           & 0.545          & 1.208          & 0.635          \\
\grey{X-Racer}                 & \grey{C + VM}                 &                                       & \grey{0.217}          & \grey{0.307 (74 \%)}  & \grey{0.747 (45 \%)}  & \multicolumn{1}{c}{\grey{---}}              \\
RoadRunner              & C + L + E              &                                       & 0.399          & 0.592          & 1.629          & 0.738          \\
Ours w/o MR             & C + VM                 &                                       & 0.241          & 0.422          & 1.261          & 0.532          \\
Ours                    & C + VM                 &                                       & \textbf{0.215} & \textbf{0.318} & \textbf{0.869} & \textbf{0.396} \\ \midrule
                        &                        & \multirow{7}{*}{\rotatebox{90}{Short Range}}          & \orange{(18.5 \%)}      & \magenta{(34.8 \%)}      & \blue{(46.7 \%)}       &                \\
LSS                     & C                      &                                       & 1.489          & 1.948          & 3.606          & 2.638          \\
Point Pillars           & L                      &                                       & 0.732          & 0.868          & 1.992          & 1.368          \\
\grey{X-Racer}                 & \grey{C + VM}                 &                                       & \grey{0.225}          & \grey{0.642 (90 \%)}  & \grey{2.421 (83 \%)}  & \multicolumn{1}{c}{\grey{---}}              \\
RoadRunner              & C + L + E              &                                       & 0.418          & 0.852          & 2.311          & 1.453          \\
Ours w/o MR             & C + VM                 &                                       & \textbf{0.276} & \textbf{0.627} & \textbf{1.835} & \textbf{1.126} \\
Ours                    & C + VM                 &                                       & 0.288          & 0.65           & 1.874          & 1.155          \\ \bottomrule
\end{tabular}
}
\end{table}

\begin{table}[t]
\centering
\renewcommand{\arraystretch}{1.3}

\caption{Evaluation of Traversability Estimation. \glsentrylong{stack} can only predict partially in (X\%) shown in gray. C: Camera, L: LiDAR, E: raw elevation, VM:Voxel Map, MR: Multi-Range}
\label{tab:wheel_risk}
\resizebox{\columnwidth}{!}{%
\begin{tabular}{@{}llcllll@{}}
\toprule
\multirow{2}{*}{Method} & \multirow{2}{*}{Input} & \multicolumn{1}{l}{\multirow{2}{*}{}} & \multicolumn{4}{c}{Risk}                                                                                      \\ \cmidrule(l){4-7} 
                        &                        & \multicolumn{1}{l}{}                       & \multicolumn{1}{c}{MSE $\downarrow$} & \multicolumn{1}{c}{Precision $\uparrow$} & \multicolumn{1}{c}{Recall $\uparrow$}   & \multicolumn{1}{c}{F1 $\uparrow$} \\ \midrule
LSS                     & C                      & \multirow{6}{*}{\rotatebox{90}{Micro Range}}                     & 0.0104                  & 0.363                         & 0.113                      & 0.173                  \\
Point Pillars           & L                      &                                            & 0.0086                  & 0.466                         & 0.189                      & 0.269                  \\
{\color[HTML]{656565}X-Racer}      & {\color[HTML]{656565}C + VM} & & {\color[HTML]{656565} 0.0056}        & {\color[HTML]{656565} 0.618 (70 \%)}  & {\color[HTML]{656565} 0.541} & {\color[HTML]{656565} 0.721}         \\
RoadRunner              & C + L + E              &                                            & 0.0086                  & 0.501                         & 0.207                      & 0.293                  \\
Ours w/o MR             & C + VM                 &                                            & 0.0080                  & 0.519                         & 0.240                      & 0.329                  \\
Ours                    & C + VM                 &                                            & \textbf{0.0076}                      & \textbf{0.523}                & \textbf{0.272}               & \textbf{0.357}                  \\ \midrule
LSS                     & C                      & \multirow{6}{*}{\rotatebox{90}{Short Range}}                     & 0.0237                  & 0.241                         & 0.241                      & 0.241                  \\
Point Pillars           & L                      &                                            & 0.0173                  & 0.431                         & 0.291                      & 0.347                  \\
{\color[HTML]{656565}X-Racer}      & {\color[HTML]{656565}C + VM} & & {\color[HTML]{656565} 0.0110}        & {\color[HTML]{656565} 0.878 (90 \%)}  & {\color[HTML]{656565} 0.537} & {\color[HTML]{656565} 0.667}\\
RoadRunner              & C + L + E              &                                            & 0.0175                  & 0.433                         & 0.305                      & 0.358                  \\
Ours w/o MR             & C + VM                 &                                            & \textbf{0.0165}                      & \textbf{0.473}                & 0.343                        & \textbf{0.397}                  \\
Ours                    & C + VM                 &                                            & 0.0166                               & 0.465                         & \textbf{0.345}               & 0.396                  \\ \bottomrule 
\end{tabular}
}
\vspace{-5ex}
\end{table}

\begin{figure*}[t]
    \centering
    \includegraphics[width=\textwidth]{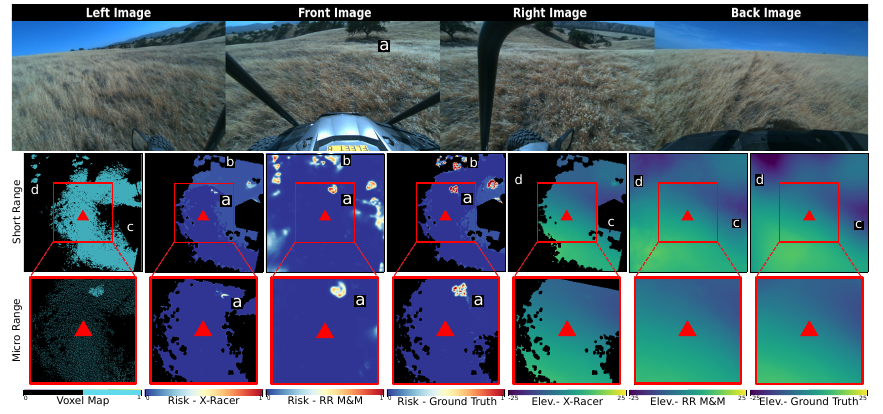}
    \caption{Qualitative results on one of the test set samples. Top: Input images, Middle: \short range maps,  Bottom: \micr range maps. The vehicle pose in the maps is shown by the red triangle. \glsentrylong{ours} is able to detect the tree (a) in front of the vehicle at \SI{45}{m}, which \glsentrylong{stack} fails to predict. \glsentrylong{stack} also fails to detect the further obstacle cluster (b) at around \SI{80}{m} which \glsentrylong{ours} is able to predict. In terms of elevation map predictions, \glsentrylong{stack} fails to predict elevation in regions missing the geometric information (c, d), while \glsentrylong{ours} is able to capture the valleys which resemble close to the ground truth elevation map.
    }
    \label{fig:visuals_1}
    \vspace{-3ex}
\end{figure*}

\subsection{Elevation Mapping Performance}
The quantitative results for elevation mapping are shown in \tabref{tab:elevation}. Compared to other camera--only (LSS), LiDAR--only (\pp) or both camera and LiDAR (RoadRunner) approaches, our approach which uses camera and voxel map as input performs the best across all regions in both \micr range and \short range. Notably, an accuracy improvement of $\sim$50\% in \micr range and  $\sim$20\% in \short range over \glsentrylong{rr} is achieved. When comparing to the \glsentrylong{stack} stack, we obtain similar performance in the \obsp regions. 
For the \obsf and \unobs regions, a direct comparison cannot be made as \glsentrylong{stack} partially estimates maps in these regions. We improved the evaluation procedure of \glsentrylong{rr} by, instead of interpolating/extrapolating the predictions of \glsentrylong{stack} to the unobserved regions, we report the coverage in percentage and performance. In the \micr range, we observe similar performance in the \obsf regions (while providing 26 \% more coverage) and slightly lower performance in the \unobs regions but provide 55\% more map coverage. For the \short range we see similar performance in the \obsf regions (with 10 \% more coverage) and significantly improved accuracy in the \unobs regions while predicting in 17\% more map regions. A qualitative result is shown in \figref{fig:visuals_1}, comparing the incomplete elevation map estimates by \glsentrylong{stack} to the map predictions of \glsentrylong{ours}. 
In addition to demonstrating improved performance, the proposed approach reduces the latency by a factor of $\sim5$ over \glsentrylong{stack}.
\par

To understand the effect of using a shared architecture for multi-range predictions, we train our approach for individual ranges separately. We observe significant accuracy improvements in the \micr range maps, especially in the \obsf (\SI{0.422}{m}$\rightarrow$ \SI{0.318}{m}) and \unobs (\SI{1.261}{m}$\rightarrow$ \SI{0.869}{m}) regions that have minimal geometric information. We hypothesize that by using a shared architecture, the multi--modal \gls{bev} features have a larger coverage ($\pm\SI{100}{m}$) and thus provide more context as compared to the individual \micr range network having a coverage of only $\pm\SI{50}{m}$. On the contrary, since the context remains the same for the \short range maps, similar performance is obtained for \short range maps in the case of the multi-range setup. This experiment highlights that a multi-range setup not only leads to overall improved performance but also avoids redundant compute for the feature extraction and fusion.



\subsection{Traversability Estimation Performance}
The quantitative results for traversability risk estimation are shown in \tabref{tab:wheel_risk}. Compared to other baselines, our approach shows the best results across all metrics at both ranges, with an improvement of up to $\sim$20\% over \glsentrylong{rr}. Moreover, we also observe improvements in the \micr range risk predictions by using the multi--range setup due to the larger context of the multi-modal \gls{bev} features, however we hypothesize that the gains are not as significant as elevation, since traversability risk is a more localized task.
\par

In comparison with \glsentrylong{stack}, our approach is able to predict in more map regions but performs slightly lower on the test set. Looking at the qualitative predictions, we observe numerous advantages of our approach over \glsentrylong{stack}. In general \glsentrylong{ours} is able to detect the obstacles from a longer range (\figref{fig:visuals_1}) while \glsentrylong{stack} is able to detect the risks only in the vicinity around the vehicle. Several instances of this can also be seen in the accompanying videos. We also note that \glsentrylong{ours} is able to reasonably detect majority of the risks but fails to precisely localize them. For example, it can associate the risk with the tree canopy fairly well; however, it fails to precisely detect the exact tree trunk location (lethal obstacle). In practice, determining the exact position of obstacles at a long distance is of less importance than detecting the presence of obstacles, given the continuous receding horizon replanning.  
Moreover, our approach is able to predict risk maps in the entire map region, even in areas without ground truth, however, this capability is not captured in the quantitative results due to lack of ground truth and their correctness can only be assessed qualitatively. \par


\begin{table}[t]
\centering
\renewcommand{\arraystretch}{1.3}
\caption{Ablation on the point cloud accumulation strategy}
\label{tab:ablation_pcl}
\resizebox{\columnwidth}{!}{%
\begin{tabular}{lcccc}
\hline
\multirow{2}{*}{Input} & \multicolumn{2}{c}{Micro Range}                                    & \multicolumn{2}{c}{Short Range}                                    \\ \cline{2-5} 
                       & \multicolumn{1}{l}{Ele. MAE [m] $\downarrow$} & \multicolumn{1}{l}{Risk F1 $\uparrow$} & \multicolumn{1}{l}{Ele. MAE [m] $\downarrow$} & \multicolumn{1}{l}{Risk F1 $\uparrow$} \\ \cline{2-5}
N=1                    & 0.592                                & 0.304                       & 1.563                                & 0.331                       \\
N=2                    & 0.521                                & 0.307                       & 1.372                                & 0.350                       \\
N=5                    & 0.466                                & 0.316                       & 1.299                                & 0.358                       \\
N=10                   & 0.438                                & 0.331                       & \textbf{1.115}                       & 0.375                       \\
VM                     & \textbf{0.403}                       & \textbf{0.358}              & 1.136                                & \textbf{0.394}              \\ \hline
\end{tabular}
}
\end{table}

\begin{table}[t]
\centering
\renewcommand{\arraystretch}{1.3}
\caption{Ablation on the loss. UL: Unobs. Loss, CL: Cons. Loss}
\label{tab:ablation_loss}
\resizebox{\columnwidth}{!}{%
\begin{tabular}{@{}lccc|ccc|c@{}}
\toprule
\multirow{2}{*}{Loss} & \multicolumn{3}{c|}{Micro Range Ele. MAE [m] $\downarrow$}                                       & \multicolumn{3}{c|}{Short Range Ele. MAE [m] $\downarrow$}                                       & \multirow{2}{*}{\begin{tabular}[c]{@{}c@{}}Cons.\\ MAE [m] $\downarrow$\end{tabular}} \\ \cmidrule{2-7}
&      \multicolumn{1}{c}{\obspt}   & \multicolumn{1}{c}{\obsft}           & \multicolumn{1}{c|}{\unobst} & \multicolumn{1}{c}{\obspt}   & \multicolumn{1}{c}{\obsft}           & \multicolumn{1}{c|}{\unobst} &                                                                              \\ \midrule
UL                                                                                                                                          & 0.229                       & 0.333                      & \textbf{0.853}                      & 0.298                       & 0.655                      & \textbf{1.849}                      & 0.18                                                                         \\
CL                                                                                                                                          & 0.216                       & 0.347                      & 1.246                      & \textbf{0.278}                       & 1.025                      & 3.575                      & \textbf{0.08}                                                                \\
UL+CL                                                                                                                                          & \textbf{0.215}              & \textbf{0.319}             & 0.870                      & 0.289                       & \textbf{0.650}             & 1.874                      & 0.09                                                                         \\ \bottomrule
\end{tabular}
}
\vspace{-2ex}
\end{table}

\begin{figure}[t]
    \centering
    \includegraphics[width=\columnwidth]{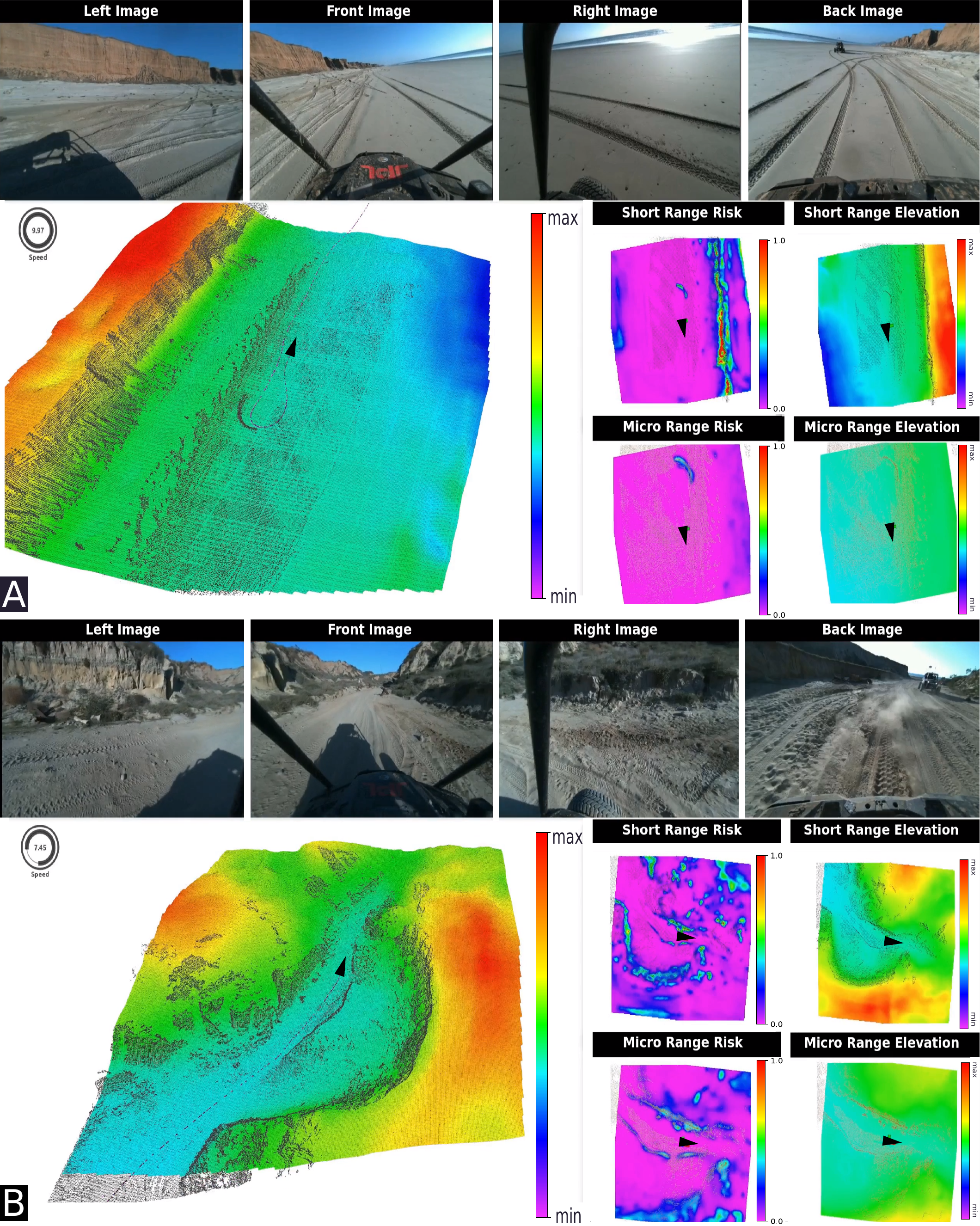}
    \caption{Predictions on various OOD environments visualized in 3D along with the top-down view of the \micr and \short range predictions. The vehicle pose is represented by a black triangle. A shows the beach environment, and B shows the canyon environment. 
    }
    \label{fig:ood}
    \vspace{-2ex}
\end{figure}

\subsection{Ablation Studies}
\label{sec:ablation}
To understand the impact of accumulated input geometric information, we vary the number of point clouds (N) accumulated temporally instead of using a voxel map as input to \glsentrylong{ours} network. We use a minimum distance threshold of \SI{2}{m} in between point clouds to avoid accumulating redundant information, and downsample the accumulated point cloud using a voxel filter of size \SI{0.4}{m}. We vary N from 1 to 10 (\tabref{tab:ablation_pcl}) and observe that less geometric information leads to worse performance in MAE and F1 score.

Next, we ablate components of the chosen loss function (\tabref{tab:ablation_loss}). First, we disable the loss in the \unobs regions ($\grid{u}{}$) by setting $\alpha = 0$. This is equivalent to not fusing the DEMs in the ground truth elevation maps and simply using the hindsight fused maps. For the \short range maps, we see a drop in performance in the \unobs regions (3.575 v.s \SI{1.874}{m}) since we are not penalizing the predictions in these regions. Interestingly, we notice that the predictions in \obsf also get worse (1.025 v.s. \SI{0.65}{m}).
We note that including the extra supervision signal in the unobserved regions is crucial and significantly improves the capability of the network to predict in regions lacking geometric information. We observe a similar trend in the \micr range elevation. Disabling the consistency loss does not largely affect the elevation MAE results but improves the consistency of elevation predictions in the overlapping regions of \micr and \short range maps (\SI{0.18}{m} $\rightarrow$ \SI{0.09}{m}). This is especially important for planning long smooth paths to facilitate high--speed navigation.

\subsection{Out--of--Distribution experiments}
We deploy our proposed approach zero--shot on out--of--distribution test datasets to evaluate its generalization performance. These environments are markedly different from the training (Paso Robles) dataset. Overall, \glsentrylong{ours} predicts accurate and consistent elevation maps (Rocky wall: \figref{fig:ood}A, canyon structure: \figref{fig:ood}B, dense forest: \figref{fig:cover_pic}) and is able to associate the corresponding traversability risks even at longer ranges. However, occasionally, it struggles to assign risk to certain unseen objects such as small Joshua trees in the Mojave desert. We also observe an interesting failure case at the San Gabriel Canyon where the network incorrectly predicts higher elevation and risk for an overhead bridge, likely due to the absence of similar overhanging structures in the training data. While we were generally surprised by the generalization capabilities to novel environments, we recommend training on a larger, more diverse dataset for improved performance. \par

\subsection{Integration with Planner}
\label{sec:planner}
\begin{figure*}[t]
    \centering
    \includegraphics[width=\textwidth]{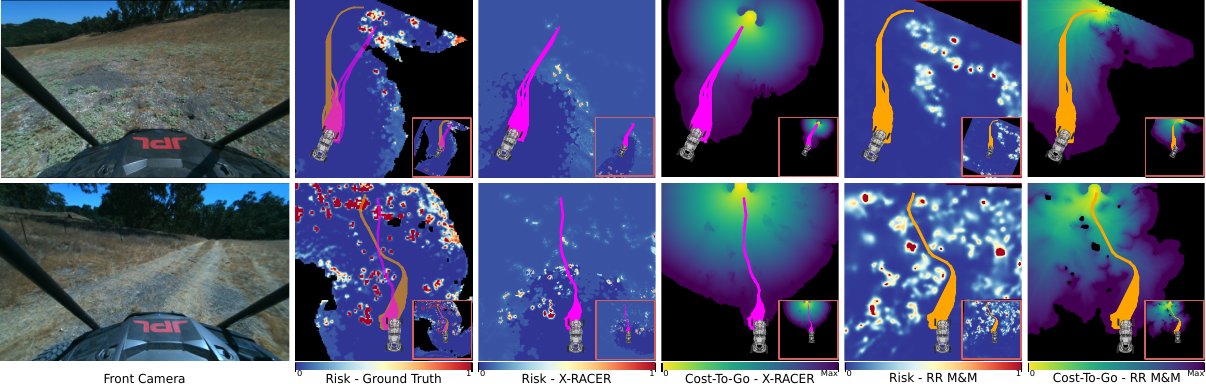}
    \caption{Qualitative results for the \short range planner. \glsentrylong{ours} is able to better predict the risks at longer ranges (resembling the ground truth risk maps) when compared to \glsentrylong{stack}. On providing a goal at a distance of \SI{100}{m} from the vehicle,  \glsentrylong{ours} planner is able to plan trajectories (orange) around the obstacles while \glsentrylong{stack} planner gives a uniform cost-to-go away from the goal point and thus plans trajectories (pink) straight through the obstacles (since it is yet to detect the obstacles at longer ranges).
    }
    \label{fig:planner_1}
    \vspace{-4ex}
\end{figure*}

The hierarchical planning stack (developed as a part of \glsentrylong{stack}) includes two stages of planning: kinematic and dynamic planning. The kinematic planner plans over a horizon of \SI{100}{m} from the vehicle (\short range) at a frequency of \SI{5}{Hz}, which is then used as input to the dynamic MPPI planner~\cite{MPPI}. The MPPI planner uses the higher resolution \micr range map and plans in the control space of steering, throttle, and brake actuator commands at a frequency of \SI{20}{Hz}. In this work, we only focus on the integration and evaluation of the \short range planner.
The kinematic lattice-based \short range planner takes into account the \short range cost maps which are queried for collision and risk values under the body and at each wheel. In addition, the slope information from the \short range elevation map constrains the maximum velocity based on slope, including constraints to avoid roll-over.
We show qualitative evaluations of the \short range planner in \figref{fig:planner_1}. In general, we observe that since \glsentrylong{ours} is able to detect the obstacles from a longer range, the planner is able to take into account these obstacles and thus plan a trajectory around them. On the contrary, \glsentrylong{stack} fails to perform predictions at longer ranges and thus leads to a uniform cost-to-go away from the goal point, causing the planner to plan a trajectory straight towards the goal, which requires pushing through dense obstacles at times. \par

\subsection{Field deployment}

We integrate \glsentrylong{ours} via a C++ ROS node for data handling, pre-processing, and map publishing, with the network implemented in Python using pybind. Inference runs on a single GPU, achieving an average time of \SI{100}{ms}, which is significantly faster than the over \SI{500}{ms} operating latency for the multi-step \glsentrylong{stack} stack. \par

We carry out an autonomous mission at Arroyo Seco, Pasadena, CA, where the vehicle navigated a \SI{400}{m} course with five waypoints. The planner stack used \short range maps from \glsentrylong{ours} and \micr range maps from \glsentrylong{stack}, allowing the vehicle to safely complete the course at speeds up to 12 m/s, successfully reaching all waypoints. For the deployment video, we refer to our webpage. Future work will focus on large-scale tests over tens of kilometers while performing comparisons with \glsentrylong{stack} in terms of predictions, path length, completion time, and number of interventions.

%% file: 05_conclusion.tex
\section{Conclusion}

In this work, we present \glsentrylong{ours}, a learning-based approach to predict traversability and elevation maps at multiple ranges for robotic off--road navigation. We demonstrate significant improvements over \glsentrylong{rr} by introducing a novel hierarchical decoder, LiDAR voxel map input, and improved supervision signal using \gls{dem}. We integrate our approach with a path planner and deploy it on real-world autonomous field experiments. While we demonstrate that the approach also generalizes to new environments, we observe that the risk predictions are not perfectly localized. Moreover, we notice that the contribution of images is relatively small compared to the voxel map in improving predictions. Future work will focus on this limitation by improving architecture for visual features and introducing temporal fusion in the \gls{bev} space. Additionally, adding uncertainty estimation for the predictions could be beneficial for the path planner.